\documentclass{article}
\usepackage{spconf,amsmath,epsfig}
\usepackage{amssymb}
\usepackage{amsmath}
\usepackage{amsfonts}
\usepackage{booktabs}
\usepackage{hyperref}

\newcommand{\cpa}[2]{\begin{minipage}[t]{ #1\columnwidth }\footnotesize{ #2}\end{minipage}}
\def\x{{\mathbf x}}
\def\y{{\mathbf y}}
\def\z{{\mathbf z}}

\def\L{{\cal L}}

\def\S{\mathcal{S}}

\title{Robust hyperspectral image classification with rejection fields}
\name{ Filipe  Condessa$~^\textbf{a,b,c,e}$, Jos{\'e} Bioucas-Dias$~^\textbf{a,b}$,  and Jelena Kova{\v c}evi{\'c}$~^\textbf{c,d,e}$\thanks{This paper was submitted to IEEE WHISPERS 2015: 7$^\textrm{th}$
Workshop on Hyperspectral Image and Signal Processing: Evolution on
Remote Sensing. The authors gratefully acknowledge support from the Portuguese Science and Technology Foundation under projects UID/EEA/50008/2013, PTDC/EEI-PRO/1470/2012, the Portuguese Science and Technology Foundation and the CMU-Portugal (ICTI) program under grant  SFRH/BD/51632/2011, NSF through award 1017278, and the CMU CIT Infrastructure Award.
}}
\address{$^\textbf{a}$Instituto de Telecomunica\c{c}\~oes, Lisboa, Portugal \\
$^\textbf{b}$Instituto Superior T\'ecnico, University of Lisbon, Lisboa, Portugal\\
$^\textbf{c}$Department of ECE, Carnegie Mellon University, Pittsburgh, PA, USA\\
$^\textbf{d}$Department of BME, Carnegie Mellon University, Pittsburgh, PA, USA\\
$^\textbf{e}$Center for Bioimage Informatics, Carnegie Mellon University, Pittsburgh, PA, USA
}
\begin{document}
%
\maketitle
\begin{abstract}
In this paper we present a novel method for robust hyperspectral image classification using context and rejection.
Hyperspectral image classification is generally  an ill-posed image problem where pixels may belong to unknown classes, and obtaining representative and complete training sets is costly.
Furthermore, the need for high classification accuracies is frequently greater than the need to classify the entire image.

We approach this problem with a robust classification method that combines classification with context with classification with rejection.
A rejection field that will guide the rejection is derived from the classification with contextual information obtained by using the SegSALSA~\cite{BioucasDiasCK:14} algorithm.
We validate our method in real hyperspectral data and show that the performance gains obtained from the rejection fields are equivalent to an increase the dimension of the training sets.
\end{abstract}
\section{Introduction}
\label{sec:intro}
Hyperspectral image classification is a challenging problem in remote sensing~\cite{bioucas2013hyperspectral}.
Due to generally ill-posed nature of hyperspectral image segmentation and classification, spatial regularization is often used (\emph{e.g.} by promoting piecewise smooth classifications) which provides context to the classification.
However, context alone cannot deal with difficulties arising from the existence of pixels belonging to unknown classes, unrepresentative and incomplete training sets, or overlapping classes.
We propose a method that, combined with contextual classification, mitigates these difficulties through the inclusion of a reject option, thus achieving robust classification.

In applications where classification performance is critical, performance gains can be obtained at the expense of not classifying all the samples.
This can be achieved by selectively abstaining from classification in situations where misclassifications are expected.
Classification with rejection was firstly analyzed in~\cite{Chow:70}, where a rejection rule for optimum error-reject trade-off was designed for binary classification.
Whereas the design of systems for classification with rejection is a rich area (see~\cite{PillaiFR:13} and references therein for state of the art systems for classification with rejection), the application of these systems is rare in pixelwise image classification and in hyperspectral image classification.

In this paper we are interested in combining classification with context with classification with rejection to obtain a robust classification scheme.
This means combining the option to reject when evidence for a classification is not enough (\emph{i.e.} reject when the classifier is likely to misclassify) with the cues that arise from spatial context information (\emph{i.e.} classification under assumption of piecewise smooth labeling). 
By associating spatial context with rejection, context cues influence the decision whether to reject or not (\emph{e.g.} a sample is less likely to be rejected if all the neighboring samples have the same label) , and rejection cues influence the context (\emph{e.g.} a sample is more likely to be rejected if all the neighboring samples are also rejected).
The robust classification idea was applied to tissue classification in stained microscopy images~\cite{CondessaBCOK_isbi:13}, where rejection is considered an extra class and Markov random fields are used as spatial contextual prior,  with significant performance improvements.
A major drawback of this approach is its rigidity with regard to the relative importance of rejection: if the amount of desired rejection is changed, the context has to be recomputed.

We propose a robust classification scheme that computes the rejection after the context,  allowing us to change amount of samples rejected on the fly.
By using the hidden fields resulting from segmentation via the constrained split augmented Lagrangian shrinkage algorithm (SegSALSA)~\cite{BioucasDiasCK:14,CondessaBDK:14}, we are able to infer a rejection field that reflects an ordering of the image pixels according to the degree of confidence associated with the contextual information, thus providing a simple and effective way to classify with rejection and context.

The paper is organized as follows:
Section \ref{sec:background} provides the background on the contextual classification algorithm (SegSALSA) and performance measures for classification with rejection.
Section \ref{sec:rej_hidden} introduces the rejection field and describes their construction and properties.
Section \ref{sec:results} presents experimental results and Section \ref{sec:conclusion} concludes the paper.

\section{Background}
\label{sec:background}
\paragraph*{SegSALSA}
The SegSALSA algorithm performs a marginal maximum \emph{a posteriori} (MMAP) segmentation through the marginalization, on the discrete labels, of a hidden field driving the probabilities~\cite{marroquin2003hidden} and applies a vectorial total variation (VTV) prior~\cite{goldluecke2012natural,sun2013classTV} on the hidden field.
This results on a convex segmentation formulation that is solved using the constraint split augmented Lagrangian shrinkage algorithm (SALSA)~\cite{afonso2011augmented}.

To describe the SegSALSA algorithm, we start by introducing notation.
Let $\x\in\mathbb{R}^{d\times n}$ represent a $n$-pixel hyperspectral image with $d$ bands and $\x_i \in \mathbb{R}^{d}$ represent the feature vector of the $i$th image pixel, with $\mathcal{S} = \{1, \hdots, n\}$ a set indexing the image pixels.
Let $\mathcal{L} = \{1, \hdots, K\}$ denote the set of possible $K$ labels, and $\y \in \mathcal{L}^n$ a labeling of the image with $y_i \in \mathcal{L}$ the label of the $i$th pixel.

Under a Bayesian perspective, the  maximum \emph{a posteriori} (MAP) labeling $\widehat{\y}$ is given by
\begin{equation}
\label{eq:map}
\widehat{\y} = \arg\max_{\y\in\mathcal{L}^n} p(\y|\x)  = \arg\max_{\y\in\mathcal{L}^n} p(\x|\y) p(\y),
\end{equation}
where $p(\y|\x)$ represents the posterior probability of the labeling $\y$ given the feature vectors $\x$, $p(\x|\y)$ the observation model, and $p(\y)$ the prior probability of the labeling $\y$.

SegSALSA approaches the segmentation, or labeling, problem by introducing a hidden field~\cite{marroquin2003hidden} $\z$  represented by a $K \times n$ matrix that, for each pixel $ i \in \S$, contains the hidden random vectors $\z_i \in \mathbb{R}^K$.
The joint probability of labels $\y$ and field $\z$ is defined as $p(\y,\z) = p(\y | \z) p(\z)$, with $p(\y | \z) = \prod_{i \in \mathcal{S}} p(y_i | \z_i)$, allowing the expression of the joint probability of the features, labels and fields $(\x,\y,\z)$ as $p(\x,\y,\z) = p(\x | \y) p(\y | \z) p(\z)$.
With the hidden field and the joint probability defined, the marginalization on the discrete labels is now possible:
\begin{equation*}
p(\x,\z) = \prod_{i\in \cal S} \big\{ \sum_{y_i \in \mathcal{L}} p(\x_i | y_i) p(y_i | \z_i) \big\} p(\z),
\end{equation*}
with the MMAP estimate being 
$\displaystyle \widehat{\z}_{\textrm{MMAP}}=\arg\hspace{-0.10in}\min_{\z \in \mathbb{R}^{K\times n}} p(\x,\z)$.

By modeling the conditional probability  $p(y_i = k|\z_i)$ as the  $k$th component of the $i$th random vector $[\z_i]_k $, two constraints are introduced in the hidden field $\z$: nonnegativity constraint ({\em i.e.}, $[\z_i]_k \geq 0$) and sum-to-one constraint ({\em i.e.}, $1^T_K z_i = 1$).
As only the discriminative power of the conditional probabilities ${\bf p}_i:=[p(\x_i | y_i=1,\dots,p(\x_i | y_i=K)]^T$ is relevant to the segmentation problem, we model them with the multinomial logistic regression (MLR) and use the logistic regression via splitting and augmented Lagrangian (LORSAL)~\cite{LiBDP:11} algorithm to learn the regression weights.

By dealing with the MMAP problem instead of the MAP, the prior is no longer applied on the discrete labels $\y$ but on the continuous hidden field $\z$.
A convex VTV prior~\cite{goldluecke2012natural,sun2013classTV} is applied on the hidden field leading to promote a smoothness along the spatial dimensions of the field, and preservation and alignment of discontinuities across the classes.

From the initial integer optimization problem in \eqref{eq:map}, the contextual classification problem is now formulated as a convex optimization problem
 \begin{align}
     \label{eq:z_optim_VTV}
    \widehat{\bf z}_{MMAP}  =  \arg\hspace{-.1in}\min_{{\bf z}\in \mathbb{R}^{K\times n} }
                                              -\sum_{i \in \S} \bigg(
                                              \ln\big({\bf p}_i^T{\bf
                                                z}_i\big) \bigg)   -\ln p( \z
                                              )  \\
                            \text{subject to:} \;\;\; {\bf z}\geq 0, \;\;\; {\bf 1}^T_K{\bf z} = {\bf 1}_n^T \nonumber.
\end{align}
Based on $\widehat{\z}_{\textrm{MMAP}}$,  $p(\y|\widehat{\z}_{\textrm{MMAP}})$ provides a soft classification,  and its maximization  with respect to $\y$ a  hard classification.
The optimization \eqref{eq:z_optim_VTV} is solved with SALSA~\cite{afonso2011augmented}, an instance of the alternating direction method of multipliers, in $O(K n \log n)$ time.
1
\paragraph*{Performance measures for classification with rejection}
To assess the performance of classification systems with rejection we use the \emph{nonrejected accuracy} $A$, the \emph{fraction of rejected samples} $r$, and the \emph{classification quality} $Q$~\cite{CondessaBK:14}.
The \emph{nonrejected accuracy} measures the accuracy on the subset of samples that are not rejected, the \emph{rejected fraction} measures how much rejection is performed, and the \emph{classification quality} jointly measures how accurate the classification on the nonrejected samples is and how inaccurate are classification on the rejected samples is.

Considering $\mathcal{S}$ the set of pixel indexes, let $\cal R$ denote the set of rejected pixels ($\cal \bar R$ the set of nonrejected samples) and $\cal C$  the set of correctly classified samples ($\cal \bar C$ the set of incorrectly classified samples).
We define the nonrejected accuracy $A$ as
\begin{equation*}
A = \frac{|\cal C \cap \cal \bar R|}{|\cal \bar R|}.
\end{equation*}
This measure, combined with the respective fraction of rejected samples, cannot compare directly the behavior of two classifiers with rejection with different rejected fractions.

The \emph{classification quality} $Q$ is defined as
\begin{equation*}
Q = \frac{|\cal C \cap \cal \bar R| + |\cal \bar C \cap \cal  R| }{|\cal S|}.
\end{equation*}
The classification quality measures the proportion of samples that are \emph{either} correctly classified and not rejected \emph{or} incorrectly classified and rejected, relative to the total number of samples.

A classifier with rejection with a classification quality of $Q$ when rejecting a fraction of samples $r$ will be equivalent, in terms of correct decisions performed, to a classifier with no rejection and accuracy numerically equal to $Q$.
The classification quality allows us to directly compare the performance of classification systems with rejection working at different rejected fractions.

\section{Rejection Field}
\label{sec:rej_hidden}
From the SegSALSA formulation and resulting hidden field we can derive a contextual rejection scheme --- the rejection field.
The hidden field $\z$ that results from the optimization problem \eqref{eq:z_optim_VTV} provides an indication of the degree of confidence associated with each label in each pixel.
This is, if $[\z_i]_k > [\z_j]_l$, we are led to believe that the label $l$ in the $j$th pixel has a smaller degree of confidence associated with the classification than the label $k$ in the $i$th pixel.

Considering the following labeling
\begin{equation*}
\widehat{\y} = \arg\max_{\y \in \L^n} p(\y| \widehat{\z}_\textrm{MMAP}),
\end{equation*}
and obtaining the associated maximum probabilities 
\begin{equation} 
\label{eq:rejfield}
[\z_{\widehat{\y}}]_i = p(\widehat{\y}_i| \widehat{\z}_\textrm{MMAP}),
\end{equation}
the probabilities associated with the MMAP labeling, we note that the same line of thought of the components of the hidden fields as an indication of confidence can be applied to the entire labeling.
If $[\z _{\widehat{\y}}]_i > [\z _{\widehat{\y}}]_j$, there is strong evidence that a higher degree of confidence exists in the labeling of the $i$th pixel as $\widehat{\y}_i$ than in the labeling of the $j$th pixel as $\widehat{\y}_j$.

We denote the $\z_{\widehat{\y}}$ field \eqref{eq:rejfield} associated with the labeling $\y$ as \emph{rejection field}.
By sorting $\z_{\widehat{\y}}$ we obtain an ordering of the samples according to their relative confidence.
The selection of a fraction of the lowest confidence samples to be rejected yields a simple, yet very effective, scheme for rejection.
This method allows not only to define, \emph{a priori}, specific values of the rejected fraction, but also to change it instantly.
Furthermore the optimal value of rejection (the rejected fraction that maximizes the classification quality) can be estimated from a subset of samples, a validation set.

The characteristics of the VTV prior used in SegSALSA indirectly impose context on the rejection field.
As it promotes smooth hidden fields, preservation of discontinuities and their alignment among classes, it preserves the discontinuities on the maximum values of the hidden field, and consequently promotes smoothness and preservation of discontinuities on the rejection field.

The computation of a rejection field and its use as a rejection rule is an approximation to the problem of contextual rejection approached in \cite{CondessaBCOK_isbi:13}, where a joint optimization on the labels and on the reject option is performed.
We perform a sequential optimization: first an optimization on the labels and then a binary optimization on the reject option through the use of a rejection field.
Whereas the solution we obtain is an approximation to the contextual rejection problem (joint minimization), the sequential optimization we perform has a clear advantage over the joint optimization approach:  the amount of rejection can be changed on the fly, whereas on the joint optimization approach the context has to be recomputed.

\section{Experimental Results}
\label{sec:results}
We illustrate the performance of our algorithm through the robust classification of the AVIRIS Indian Pine scene, and the ROSIS Pavia university scene.
The Indian Pine scene was acquired with the AVIRIS sensor in NorthWest Indiana (USA), being a $145 \times 145$ pixel hyperspectral image with $200$ spectral bands (excluding water absorption bands) containing $16$ not mutually exclusive classes.
The Pavia University scene was acquired with the ROSIS sensor in Pavia (Italy), being a $610 \times 340$ pixel hyperspectral image with $103$ spectral bands  containing $9$ not mutually exclusive classes.
We model the MLR weights with LORSAL and use the SegSALSA algorithm to include context in the classification.
\def\wi{0.39}
\def\hi{0.4}
\begin{figure*}[ht]
\begin{center}
\begin{tabular}{ccccc}
\includegraphics[width=\wi\columnwidth,height=\hi\columnwidth]{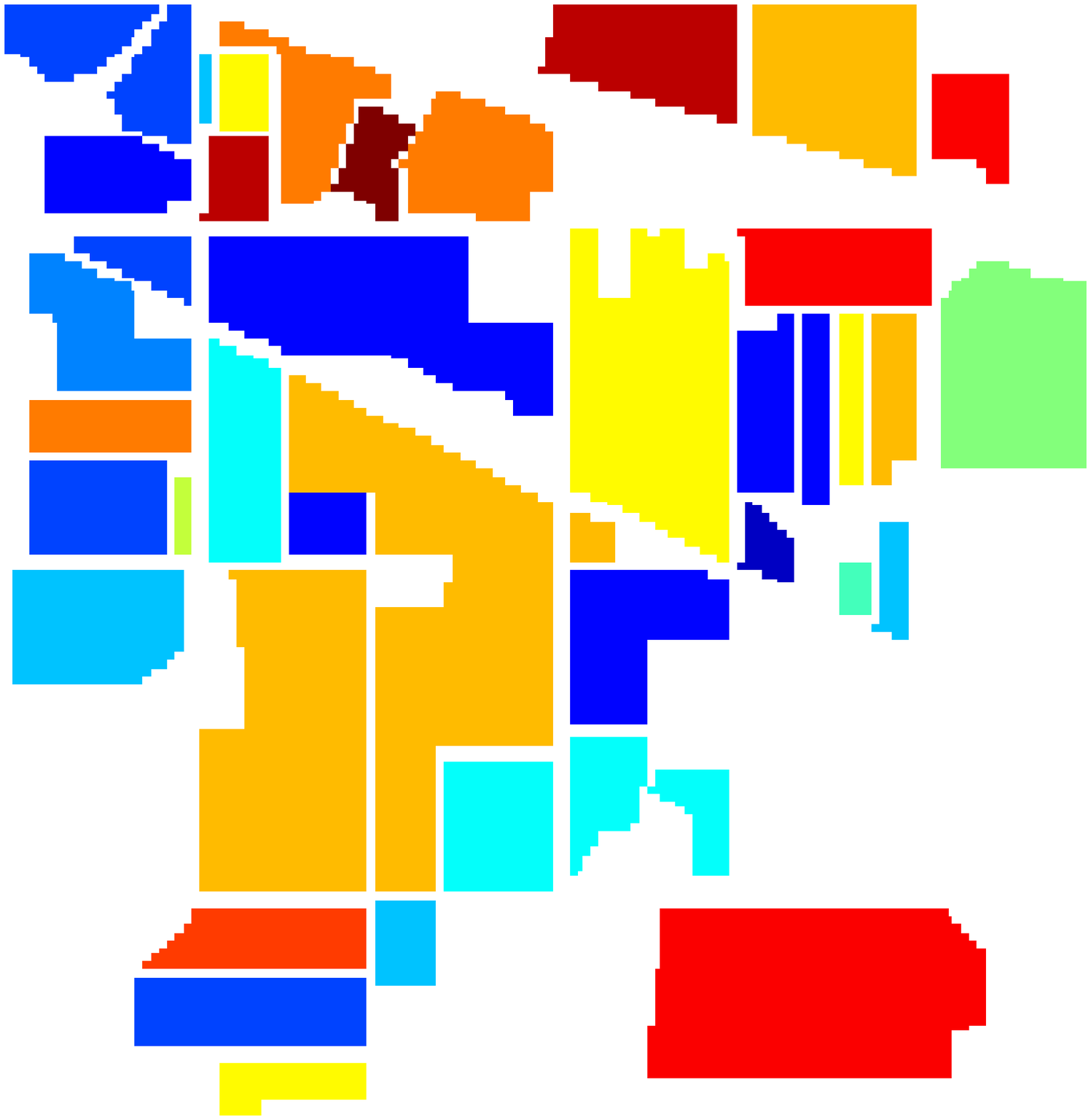} &
\includegraphics[width=\wi\columnwidth,height=\hi\columnwidth]{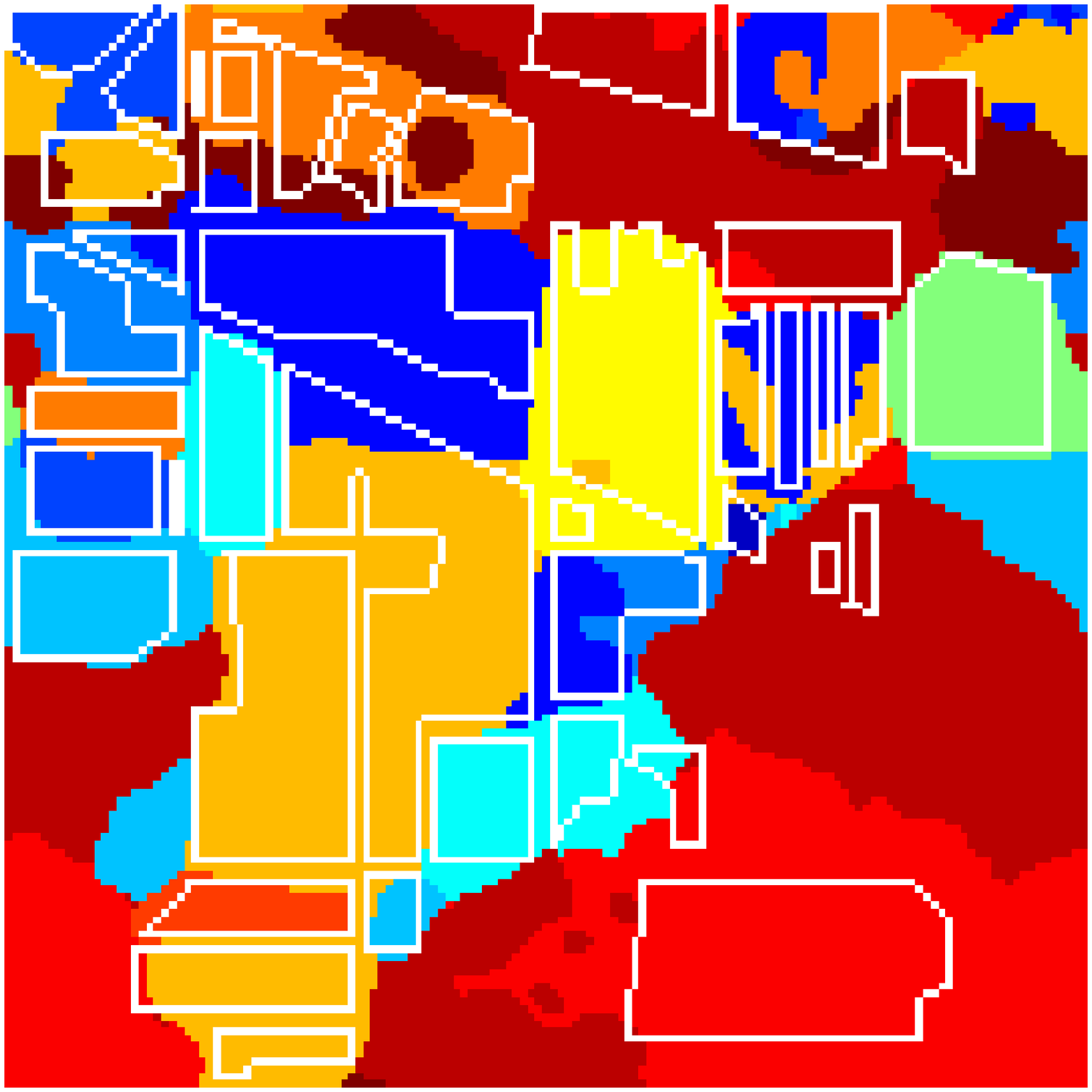} &
\includegraphics[width=\wi\columnwidth,height=\hi\columnwidth]{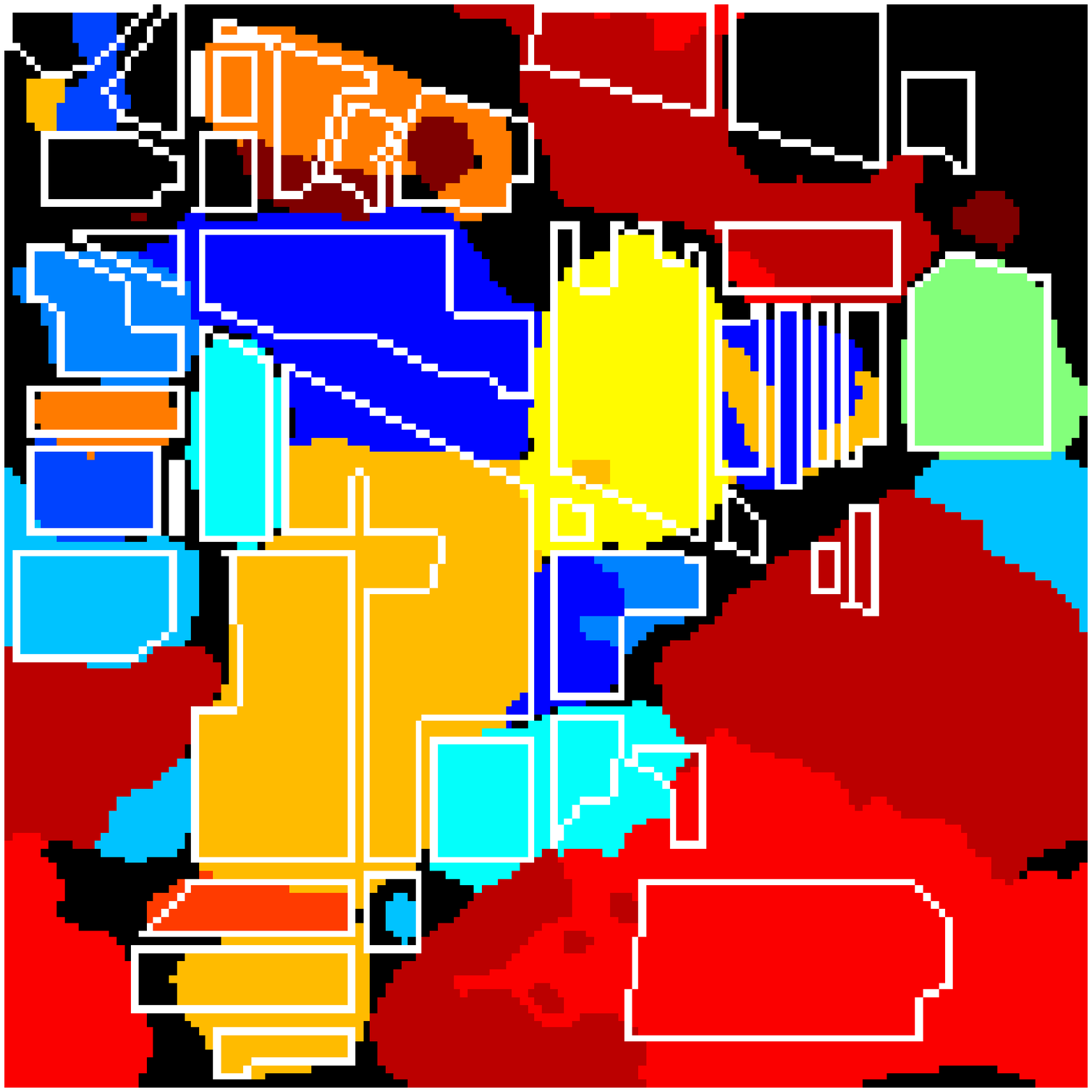} &
\includegraphics[width=\wi\columnwidth,height=\hi\columnwidth]{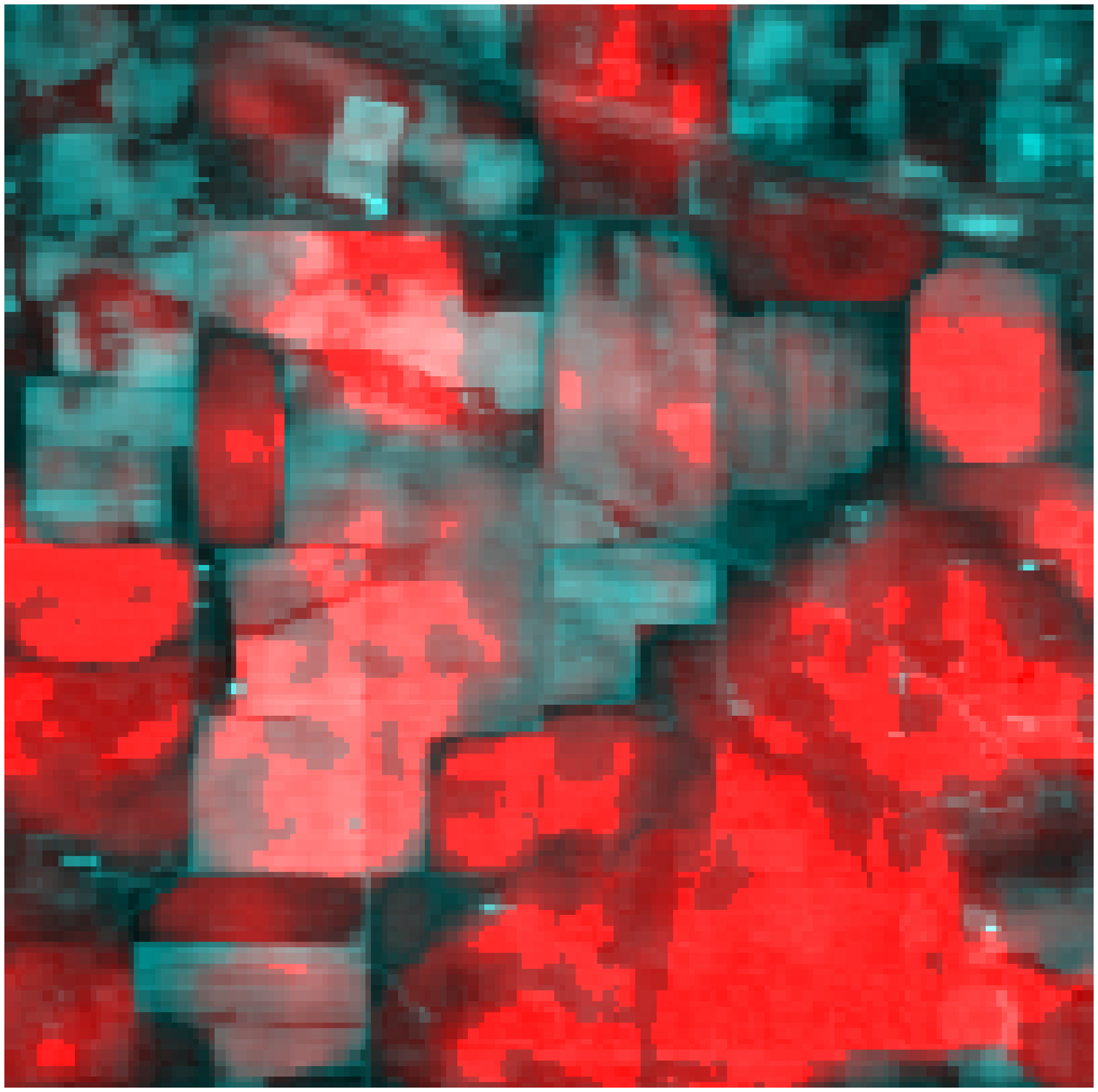} &
\includegraphics[width=\wi\columnwidth,height=\hi\columnwidth]{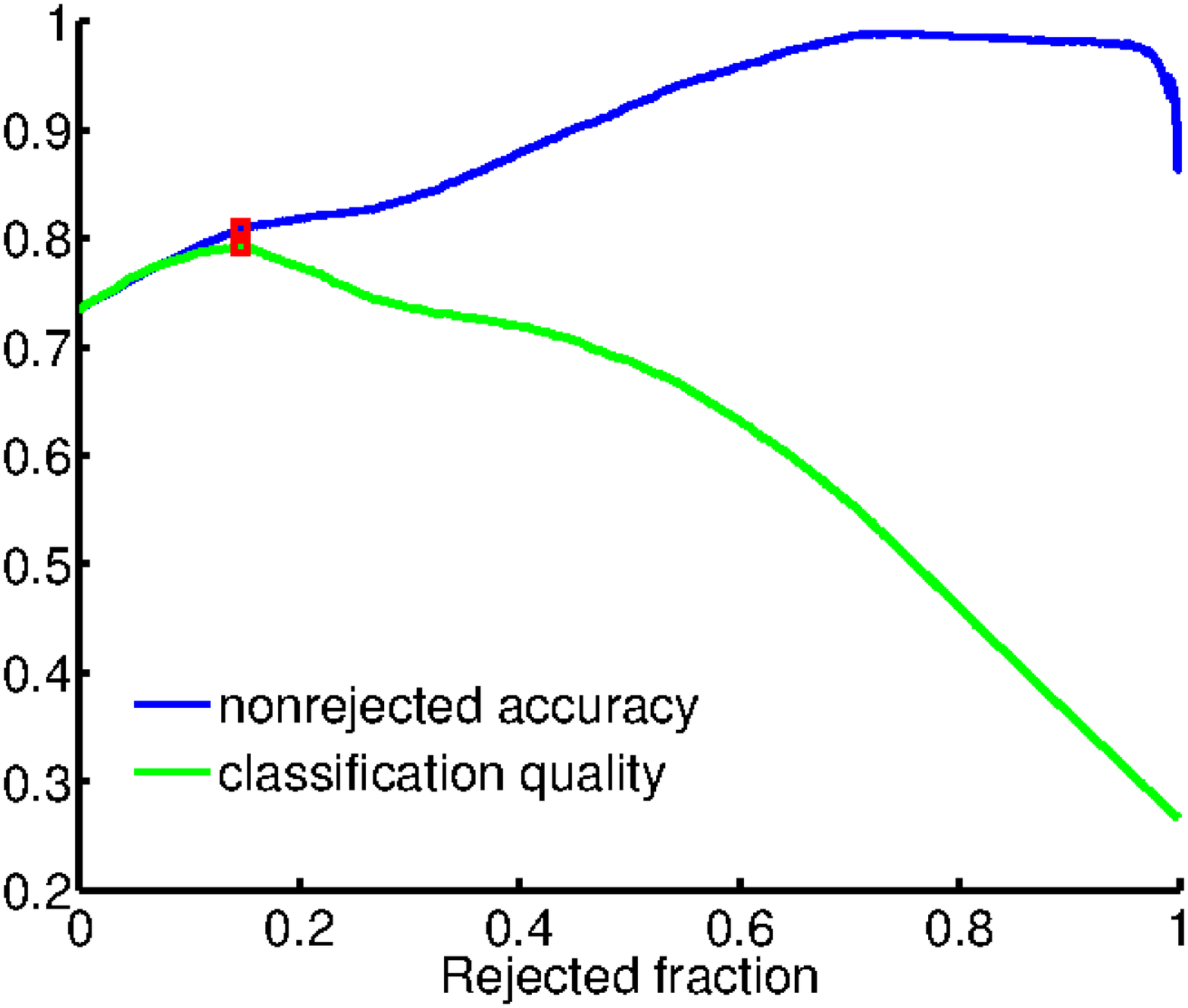}\\
(a) & (b) & (c) & (d) &(e) \\
\includegraphics[width=\hi\columnwidth,height=\wi\columnwidth, angle =90]{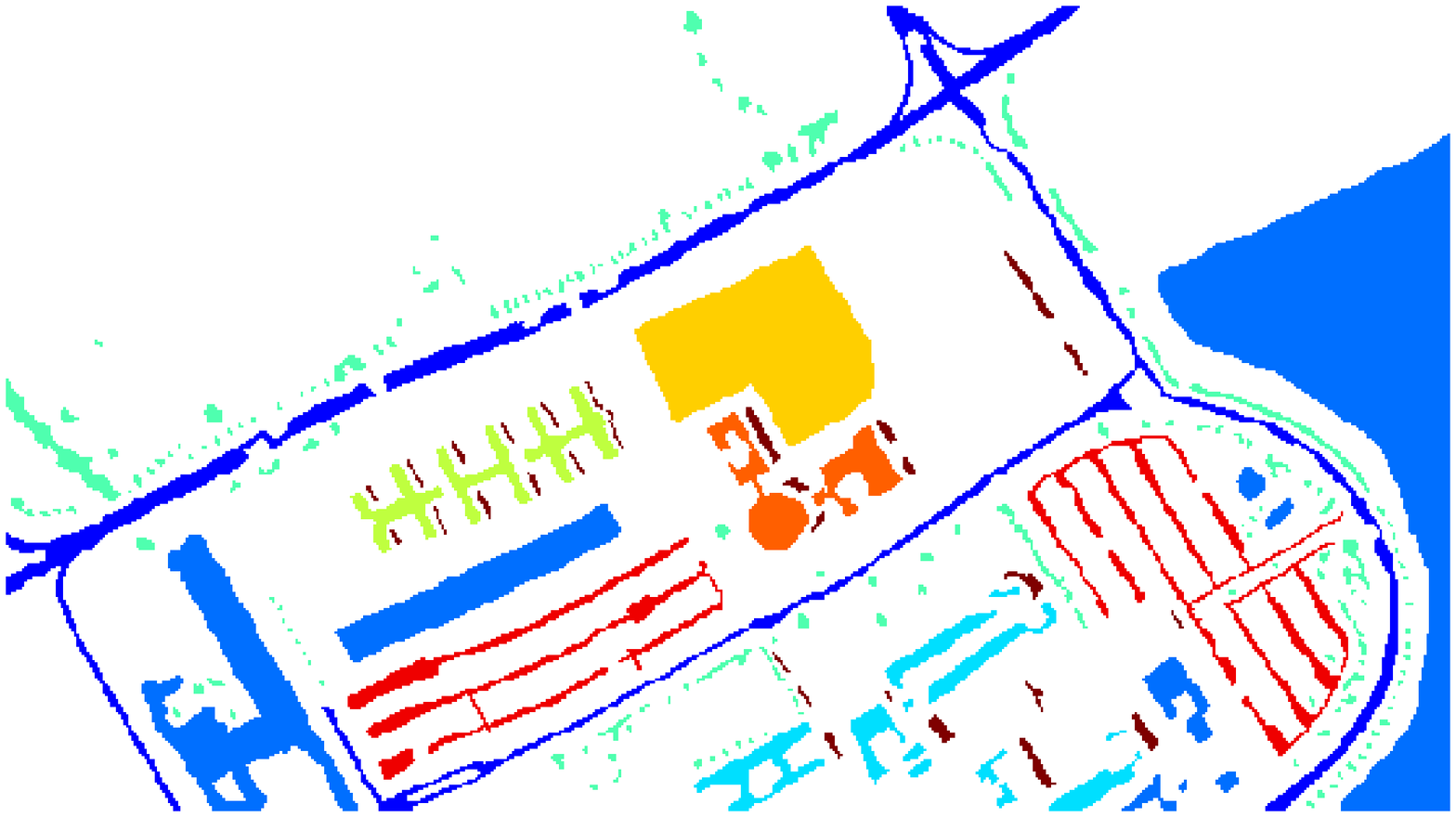} &
\includegraphics[width=\hi\columnwidth,height=\wi\columnwidth, angle =90]{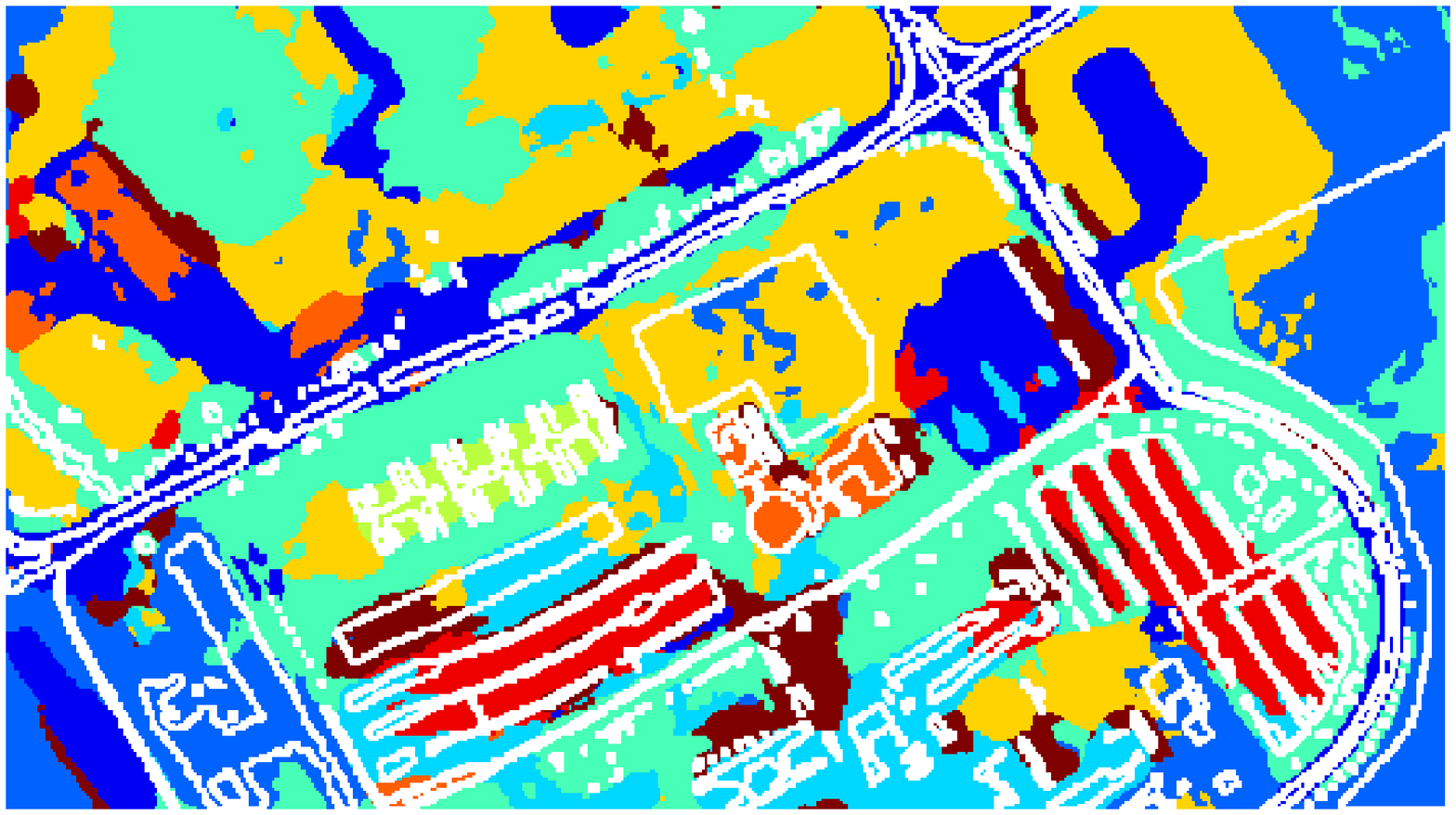} &
\includegraphics[width=\hi\columnwidth,height=\wi\columnwidth, angle =90]{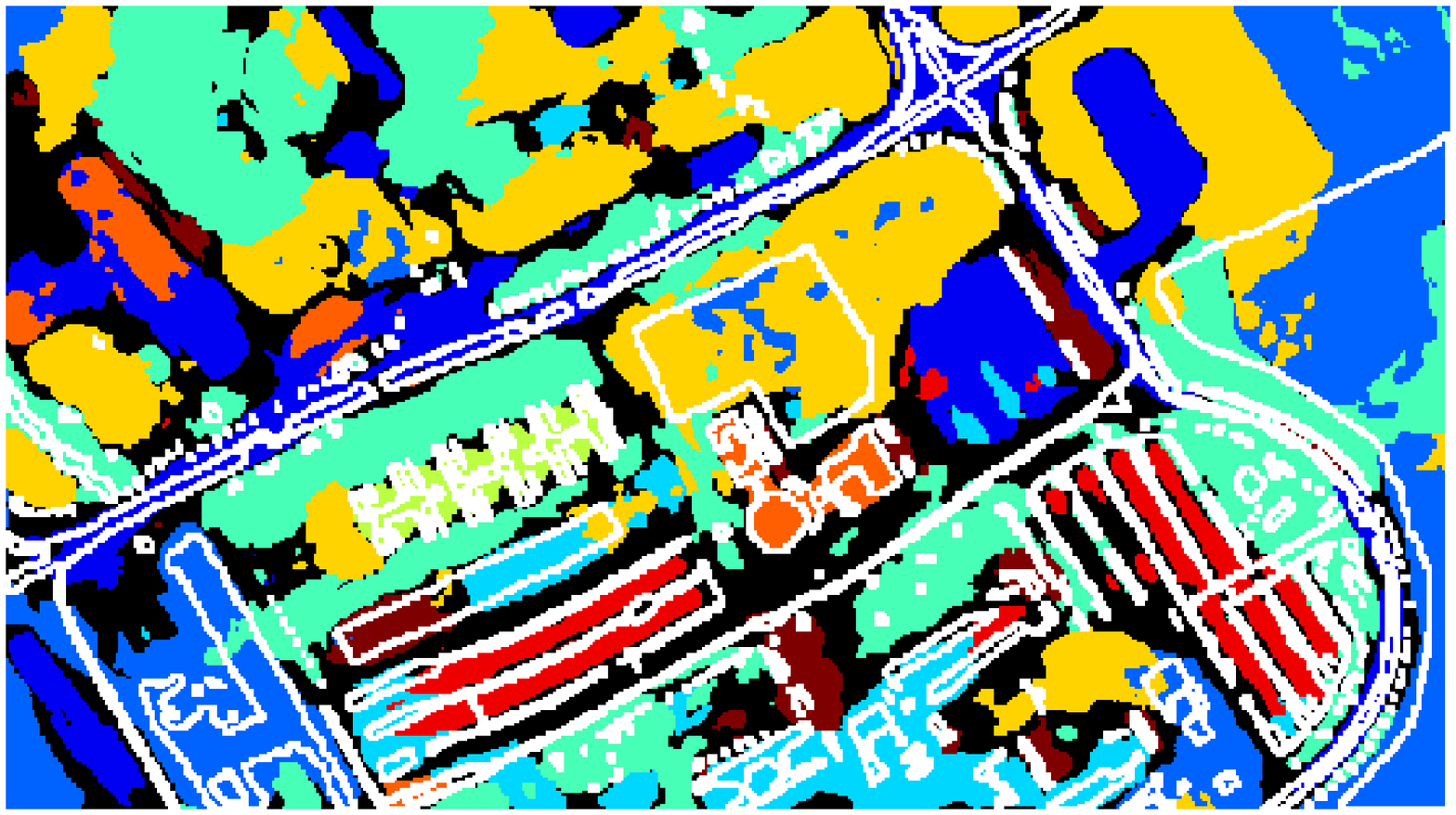} &
\includegraphics[width=\hi\columnwidth,height=\wi\columnwidth, angle =90]{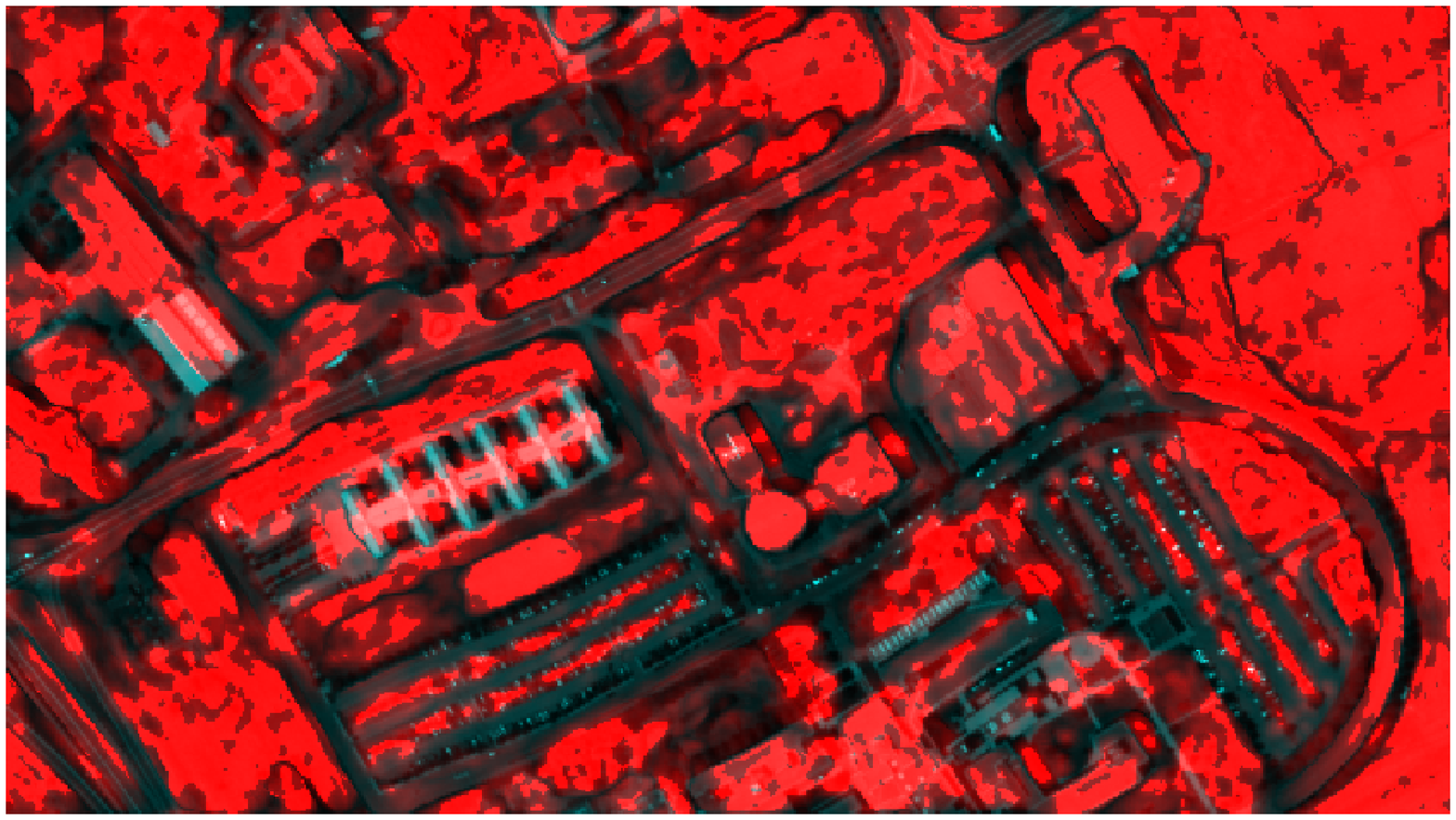} &
\includegraphics[width=\wi\columnwidth,height=\hi\columnwidth]{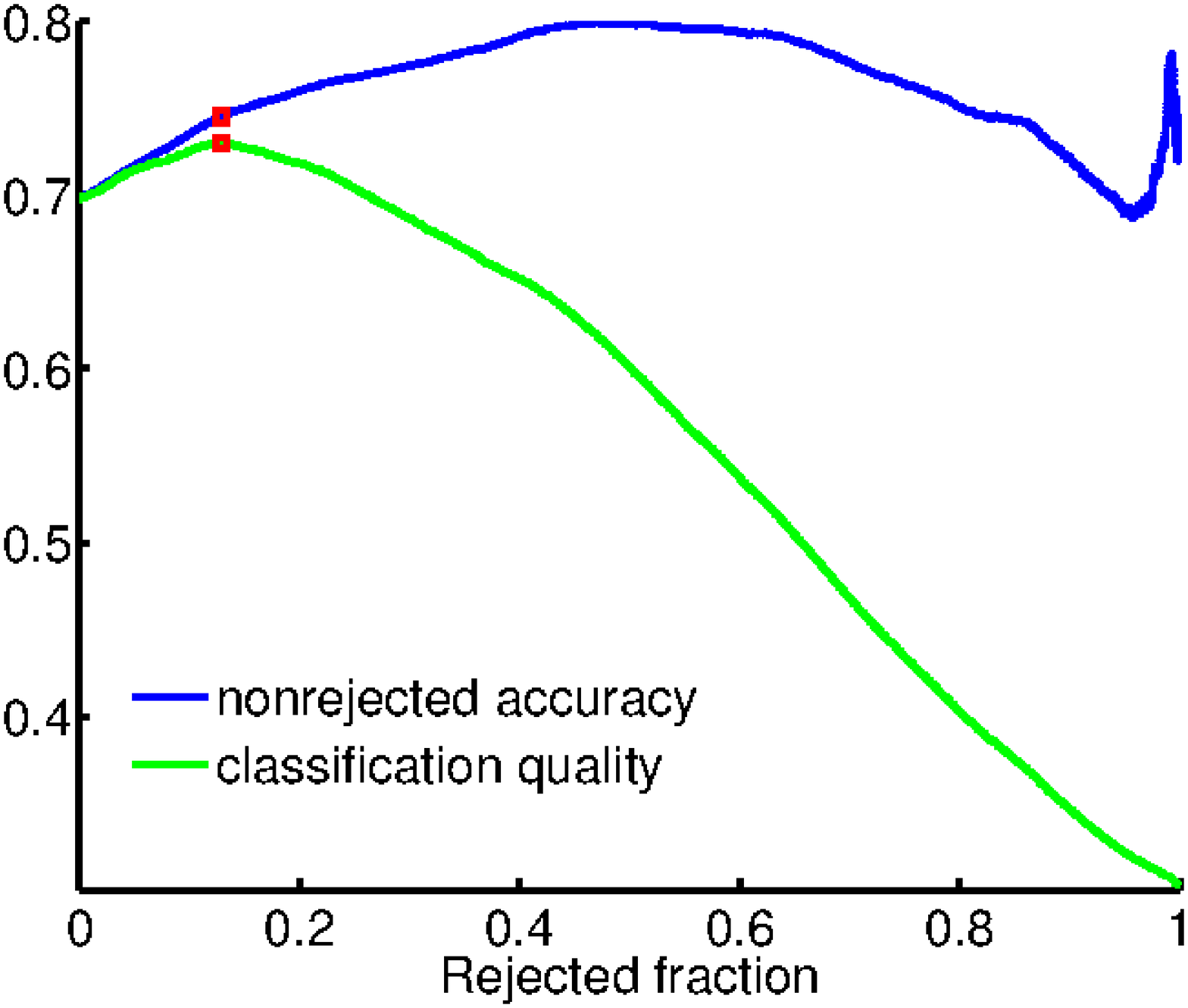}\\
(f) & (g) & (h) & (i) & (j) \\

\end{tabular}
\caption{ \label{fig:results} Top row: Robust classification of the Indian Pines scene. (a) Ground truth  and (b) classification with $15$ training samples per class using LORSAL and SegSALSA ($73.5\%$ accuracy), (c) classification with optimal rejected fraction ($80.9\%$ nonrejected accuracy at a rejected fraction of $14.7\%$ with classification quality of $79.2\%$), and (d) associated rejection fields. (e) Nonrejected accuracy and classification quality variation with rejected fraction (maximum classification quality in red).
Bottom row: Robust classification of the Pavia University scene. (f) Ground truth  and (g) classification with $15$ training samples per class using LORSAL and SegSALSA ($69.8\%$ accuracy), (h) classification with optimal rejected fraction ($74.6\%$ nonrejected accuracy at a rejected fraction of $12.9\%$ with classification quality of $73.0\%$), and (i) associated rejection fields. (j) Nonrejected accuracy and classification quality variation with rejected fraction (maximum classification quality in red).}
\end{center}
\end{figure*}

\begin{table}
 {\caption{\label{tb:classwise} Classwise performance measures for classification with rejection of the Indian Pines scene (Fig.  \ref{fig:results}, top row).
$OA$ correspondes to the accuracy of the SegSALSA classification method with no rejection (Fig. \ref{fig:results} b), and $A$ corresponds to nonrejected accuracy, $Q$ to classification quality, and $r$ to rejected fraction from classification with rejection (Fig. \ref{fig:results} c). $n$ Is the number of samples per class.
}}
{\small
 \begin{tabular}{lrrrrr}
& $OA$ $(\%)$ & $A (\%)$ &  $Q (\%)$& $r (\%)$ & $n$ \\
\midrule
alfafa &$71.74$ &$0.00$ &$26.09$ &$\bf 97.83$ &$46$ \\
corn no-till &$66.67$ &$76.57$ &$ \bf 79.06$ &$13.94$ &$1428$ \\
corn min-till &$53.13$ &$47.33$ &$43.49$ &$36.87$ &$830$ \\
corn clean &$100.00$ &$100.00$ &$96.62$ &$3.38$ &$237$ \\
grass past.&$77.85$ &$81.06$ &$75.78$ &$13.66$ &$483$ \\
grass trees&$90.55$ &$90.83$ &$90.00$ &$1.37$ &$730$ \\
grass mowed&$0.00$ &$0.00$ &$0.00$ &$0.00$ &$28$ \\
hay&$99.16$ &$100.00$ &$97.70$ &$3.14$ &$478$ \\
oats&$0.00$ &$0.00$ &$100.00$ &$100.00$ &$20$ \\
soybean no-till&$72.94$ &$74.09$ &$71.81$ &$7.10$ &$972$ \\
soybean min-till &$72.38$ &$88.54$ &$\bf 89.53$ &$19.67$ &$2455$ \\
soybean clean&$79.26$ &$78.50$ &$69.48$ &$14.50$ &$593$ \\
wheat &$86.34$ &$86.21$ &$85.37$ &$0.98$ &$205$ \\
woods &$74.55$ &$81.56$ &$82.29$ &$9.96$ &$1265$ \\
bldg. &$66.06$ &$80.70$ &$84.20$ &$18.13$ &$386$ \\
stone &$32.26$ &$32.26$ &$32.26$ &$0.00$ &$93$ \\
 \bottomrule
 \end{tabular}}
\end{table}
Figure \ref{fig:results} illustrates the performance gains obtained by combining classification with context with classification with rejection.
Using the rejection field, we are able to change the amount of rejected samples on the fly, without need to recompute the context.
Table \ref{tb:classwise} shows that the performance gains are not equally distributed among all classes. The bulk of the performance gains is achieved by increasing the performance in highly populated classes.
This is achieved either by a minor drop in nonrejected accuracy in small number of lesser populated classes, or by the entire rejection of lesser populated class.

The performance gains obtained from the allocation of labeled samples to estimate the optimal rejected fraction (the rejected fraction that maximizes the classification quality) can be larger than the gains obtained from using those samples to extend the training set, retraining with LORSAL and classifying the image with SegSALSA.
This effect is clearly illustrated on table \ref{tb:table}, where, in the Indian Pines scene, for an initial training set of $30$ samples the class, the effect of either estimating the optimal rejected fraction from $50$ randomly selected samples or retraining the classifier with the extra $50$ samples is shown.
\begin{table}[h]
\begin{center}
\caption{\label{tb:table} Effect of increasing the dimension of the training set with new samples \emph{vs}. using the new samples as validation set to estimate the rejected fraction $r$ in the Indian Pines scene. Comparison of average performance (classification quality $Q$, nonrejected accuracy $A$, and rejected fraction $r$) over $30$ Monte Carlo runs.}
\begin{tabular}{lccc}
 & \small $r$ ($\%$) &\small $Q$ ($\%$) &\small $A$ ($\%$) \\
\midrule
\hspace{-.1in}\cpa{.55}{\textbf{initial} -- training set of $480$ samples with no rejection}& \small $0.00$ &\small$84.21 $  &\small$84.21 $  \\ 
\hspace{-.1in}\cpa{.55}{\textbf{extended} -- training set of $480 + 50$ samples with no rejection} &\small$0.00$ &\small $ 86.46$ & \small $ 86.46$ \\
\hspace{-.1in}\cpa{.55}{\textbf{estimated} -- training set of $480$ samples, with optimal rejected fraction estimated from $50$ samples} &\small $12.77$ &\small $ \textbf{87.02} $ &\small $ 91.16 $ \\
\hspace{-.1in}\cpa{0.55}{\textbf{optimal} -- training set of $480$ samples, with true optimal rejected fraction}& \small$12.49$ &\small $88.37$ & \small $ 91.53 $ \\
\bottomrule
\end{tabular}
\end{center}
\end{table}
Whereas it is clear that the increased performance obtained by estimating the rejected fraction when compared to retraining the classifier will not hold for smaller training sets, for larger training sets it is a computationally cheaper and performance-wise better alternative to retraining the classifier.

\section{Concluding Remarks}

\label{sec:conclusion}
We presented a simple and effective scheme for robust hyperspectral image classification by combining classification with context and classification with rejection by deriving a rejection field from the hidden fields that drive the contextual classification.
We moved from the joint optimization problem of context and rejection, to a faster separate optimization without losing the contextual effect on the rejection.
The performance gains obtained by using robust classification are shown to be equivalent to training the classifier with larger training sets.

\paragraph*{Acknowledgements}
The authors would like to thank D. Landgrebe at Purdue University for providing the AVIRIS Indian Pines scene, P. Gamba at Pavia University for providing the ROSIS Pavia University scene.

\bibliography{}
\bibliographystyle{IEEEbib}
\end{document}